%% file: paper_v3.tex
\documentclass[conference]{IEEEtran}
\IEEEoverridecommandlockouts
\usepackage{cite}
\usepackage{amsmath,amssymb,amsfonts}
\usepackage{algorithmic}
\usepackage{graphicx}
\usepackage{textcomp}
\usepackage{mathtools}
\usepackage{xcolor}
\def\BibTeX{{\rm B\kern-.05em{\sc i\kern-.025em b}\kern-.08em
    T\kern-.1667em\lower.7ex\hbox{E}\kern-.125emX}}
\usepackage{url}

\usepackage{tikz}
    \usetikzlibrary{shapes.arrows}
    
\usepackage{pgfplots}
\pgfplotsset{
compat=1.3,
legend style={font=\footnotesize, fill opacity=0.7,  draw opacity=1, text opacity=1, draw=white!15!black, legend cell align=left, align=left}, 
width=6cm, 
height=6cm,
yminorticks=false,
xminorticks=false,
title style={font=\small},
tick style={color=black},
tick label style={font=\small},
grid style={line width=.1pt, draw=gray!20},
major grid style={line width=.1pt,draw=gray!20},
}
\usepgfplotslibrary{colorbrewer}
\usepgfplotslibrary{groupplots}
\usepgfplotslibrary{fillbetween}
\usetikzlibrary{plotmarks}
\usetikzlibrary{patterns}
\pgfplotsset{every tick label/.append style={font=\footnotesize}}

\usepackage{subfig}

\usepackage{glossaries}
\input{acronyms.tex}

\newcommand{\E}[1]{\mathbb{E}\left[ #1 \right]} % expectation
\newcommand{\mc}[1]{\mathcal{#1}}   % mathcal abbreviation
\DeclareMathOperator*{\argmax}{arg\,max}    % argmax
\DeclareMathOperator*{\argmin}{arg\,min}    % argmin

\def \fwidth{0.32\linewidth}
\def \fheight {0.2\linewidth}

\def \sfwidth{0.48\columnwidth}
\def \sfheight {0.32\columnwidth}

\definecolor{color4}{HTML}{FFD700}
\definecolor{color3}{HTML}{EA5F94}
\definecolor{color2}{HTML}{CD34B5}
\definecolor{color1}{HTML}{9D02D7}
\definecolor{color0}{HTML}{0000FF}
\definecolor{darkblue}{HTML}{00429D}
\definecolor{darkgreen}{HTML}{005c00}
\definecolor{gold}{HTML}{D4AF37}
\definecolor{darkred}{HTML}{910000}
\definecolor{darkslategray38}{RGB}{38,38,38}

\begin{document}

\title{Semantic and Effective Communication for Remote Control Tasks with Dynamic Feature Compression}

\author{\IEEEauthorblockN{Pietro Talli, Francesco Pase, Federico Chiariotti, Andrea Zanella, Michele Zorzi}
\IEEEauthorblockA{Department of Information Engineering, University of Padova, Via G. Gradenigo 6/B, 35131 Padua, Italy \\
Emails: \{pietro.talli, francesco.pase\}@studenti.unipd.it, \{federico.chiariotti, andrea.zanella, michele.zorzi\}@unipd.it\vspace{-0.7cm}}
\thanks{This work was supported by the Italian Ministry of University and Research under the PNRR ``SoE Young Researchers'' grant for project REDIAL.}
}

\maketitle

\begin{abstract}
The coordination of robotic swarms and the remote wireless control of industrial systems are among the major use cases for 5G and beyond systems: in these cases, the massive amounts of sensory information that needs to be shared over the wireless medium can overload even high-capacity connections. Consequently, solving the \emph{effective communication} problem by optimizing the transmission strategy to discard irrelevant information can provide a significant advantage, but is often a very complex task. In this work, we consider a prototypal system in which an observer must communicate its sensory data to an actor controlling a task (e.g., a mobile robot in a factory). We then model it as a remote \gls{pomdp}, considering the effect of adopting semantic and effective communication-oriented solutions on the overall system performance. We split the communication problem by considering an ensemble \gls{vqvae} encoding, and train a \gls{drl} agent to dynamically adapt the quantization level, considering both the current state of the environment and the memory of past messages. We tested the proposed approach on the well-known CartPole reference control problem, obtaining a significant performance increase over traditional approaches.
\end{abstract}


\begin{tikzpicture}[remember picture,overlay]
	\node[anchor=north,yshift=-25pt] at (current page.north) {\parbox{\dimexpr\textwidth-\fboxsep-\fboxrule\relax}{
			\centering\footnotesize This paper has been submitted to IEEE for publication. Copyright may change without notice. \\
	}};
\end{tikzpicture}

\glsresetall

\section{Introduction}
\label{sec:intro}

In the introduction to Shannon's seminal work on communication theory~\cite{shannon1949mathematical}, Warren Weaver defined a problem with three levels: classical information theory deals with Level A, or the \emph{technical problem}, which concerns itself with the accurate and efficient transmission of data. However, if communication is oriented toward affecting an agent's action, there are two further levels to the problem: Level B, or the \emph{semantic problem}, is to find the most effective way to convey the meaning of the message, even when irrelevant details are lost or misunderstood, while Level C, also called the \emph{effectiveness problem}, deals with the resulting behavior of the receiver: as long as the receiver takes the optimal decision, the effectiveness problem is solved, regardless of any differences in the received information.

While the Level B and C problems attracted limited attention for decades, the explosion of \gls{iiot} systems has drawn the research and industrial communities toward semantic and effective communication~\cite{popovski2020semantic}, optimizing remote control processes under severe communication constraints beyond Shannon's limits on Level A performance~\cite{gunduz2022beyond}. In particular, the effectiveness problem is highly relevant to robotic applications, in which independent mobile robots, such as drones or rovers, must operate based on information from remote sensors. In this case the sensors and cameras act as the transmitter in a communication problem, while the robot is the receiver: by solving the Level C problem, the sensors can transmit the information that best directs the robot's actions toward the optimal policy~\cite{stavrou2022rate}. We can also consider a case in which the robot is the transmitter, while the receiver is a remote controller, which must get the most relevant information to decide the control policy~\cite{wan2020cognitive}.

The rise of communication metrics that take the content of the message into account, such as the \gls{voi}~\cite{yates2020agesurvey}, represents an attempt to approach the problem in practical scenarios, and analytical studies have exploited information theory to define a semantic accuracy metric and minimize distortion~\cite{shao2022theory}. In particular, \emph{information bottleneck} theory~\cite{beck2022semantic} has been widely used to characterize Level B optimization~\cite{shao2021learning}. However, translating a practical system model into a semantic space is a non-trivial issue, and the semantic problem is a subject of active research~\cite{uysal2022semantic}.
The effectiveness problem is even more complex, as it implicitly depends on estimating the effect of communication distortion on the control policy and, consequently, on its performance~\cite{tung2021effective}. While the effect of simple scheduling policies is relatively easy to compute~\cite{kim2019learning}, and linear control systems can be optimized explicitly~\cite{zheng2020urgency}, realistic control tasks are highly complex, complicating an analytical approach to the Level C problem. Pure learning-based solutions that consider communication as an action in a multi-agent \gls{drl} problem, such as emergent communication, also have limitations~\cite{foerster2016learning}, as they can only deal with very simple scenarios due to significant convergence and training issues. In some cases, the information bottleneck approach can also~\cite{goyal2018transfer} be exploited to determine state importance, but the existing literature on optimizing Level C communication is very sparse, and limited to simpler scenarios~\cite{rcmab_pase}.

In this work, we consider a dual model which combines concepts from \gls{drl} and semantic source coding: we consider an ensemble \gls{vqvae} model~\cite{van2017neural}, each of which learns to represent observations using a different codebook. A \gls{drl} agent can then select the codebook to be used for each transmission, controlling the trade-off between accuracy and compression. Depending on the task of the receiver, the reward to the \gls{drl} agent can be tuned to solve the Level A, B, and C problems, optimizing the performance for each specific task.
In order to test the performance of the proposed framework, we consider the well-known CartPole problem, whose state can be easily converted into a semantic space, as its dynamics depend on a limited set of physical quantities. We show that dynamic codebook selection outperforms static strategies for all three levels, and that considering the Level C task can significantly improve the control performance with the same bitrate.

The rest of the paper is organized as follows: first, we present the general system model in Sec.~\ref{sec:system}. We then present the dynamic feature compression solution in Sec.~\ref{sec:solution}, which is evaluated by simulation in Sec.~\ref{sec:results}. Finally, Sec.~\ref{sec:conclusion} concludes the paper, along with remarks on possible extensions.

\section{System Model}
\label{sec:system}

We consider a general model in which two agents cooperate to solve a \emph{remote \gls{pomdp}} problem. In the following, we will denote random variables with capital letters, their possible values with lower-case letters, and sets with calligraphic or Greek capitals.

In the standard \gls{pomdp} formulation~\cite{pomdp}, one agent needs to optimally control a stochastic process defined by a tuple $\left< \Sigma, \mathcal{A}, \Omega, P, \xi, R, \gamma\right>$, where $\Sigma$ represents the set of system states, $\mathcal{A}$ is the set of feasible actions, and $\Omega$ is the observation set. The function $P : \Sigma \times \mathcal{A} \rightarrow \Phi(\Sigma)$, where $\Phi(\cdot)$ represents all possible probability distributions over a set, gives the state transition probability function, with $P_{ss'}^a$ denoting the conditional probability of entering state $s' \in \Sigma$ given the current state $s$ and the taken action $a \in \mathcal{A}$. On the other hand, $\xi : \Sigma \rightarrow \Phi(\Omega)$ is an observation function, which provides the conditional probabilities %$\xi(o|s) = \mathrm{Pr} \left[ S_t = s | O_t = o\right]$
$\xi_{os}$, i.e., the probability that the system is in state $s$, given the observation $o$. Finally, $R: \Sigma \times \mathcal{A} \rightarrow \mathbb{R}$ provides the average reward given to the agent when taking action $a$ in state $s$, denoted as $R_s^a$, and the scalar $\gamma \in [0, 1)$ is a discount factor.

Specifically, the \gls{pomdp} proceeds in discrete steps indexed by $t$, and at each step $t$ the agent can infer the system state $s_t$ only from the stochastic observation $o_t$, thus having partial information on it. Based on this observation, and on its policy $\pi : \Omega \rightarrow \Phi(\mathcal{A})$, which outputs a probability distribution over the action space for each possible observation, the agent interacts with the system by sampling an action ${a_t \sim \pi(a_t | o_t)}$. Then, given the real system state $s_t$ and the sampled action $a_t$, the agent receives a feedback from the system in the form of a reward ${r_t = R_{s_t}^{a_t}}$. The goal for the agent is to optimize its policy $\pi$ to maximize the expected cumulative discounted reward $G = \mathbb{E}\Big[ \sum_{t} \gamma^t R_t \Big]$. We can see that the \gls{pomdp} is indeed built on top of the \gls{mdp} formulation, with the addition of noisy observations.

\subsection{The Remote POMDP}

In this paper we consider a variant of the \gls{pomdp}, that we define \emph{remote \gls{pomdp}}, in which two agents are involved in the process. The first agent,  i.e., the \emph{observer}, receives the observation $o_t \in \Omega$, and needs to convey such information to a second agent, i.e., the \emph{actor}, through a constrained communication channel, which limits the number of bits the observer can send. Consequently, the amount of information the observer can send to the actor is limited. The actor then chooses and takes an action. This system can formalize many control problems in future \gls{iiot} systems, as sensors and actuators can potentially be geographically distributed, and the amount of information they can exchange to accomplish a task is limited due to the shared wireless medium, which has to be allocated to the many devices installed in the factory, as well as by the energy limitations on the sensors.

% ------------------------- %

In order to transmit the most relevant information to the actor, the observer chooses its transmission $m_t \in \mathcal{M}$, where $\mathcal{M}$ is the set of possible messages (which corresponds to the observer's action space). Then, based on the received message $m_t$, the actor optimizes its own policy $\pi^a(a_t | m_t)$ to maximize $G$, and to optimally solve the underlying control problem. As the objective of the observer is to minimize channel usage, i.e., communicate as few bits as possible, while maintaining the highest possible performance in the control task, we consider the problem of adaptive encoding.

In order to solve this problem, the observer needs to have a way to gauge the \emph{value} of information, as described in the introduction: information theory, and in particular rate-distortion theory, have provided the fundamental limits when optimizing for the technical problem, i.e., Level A, where the goal is to reconstruct the source signals with the highest fidelity~\cite{cover:IT}. The first and simplest way to solve the remote \gls{pomdp} problem is to blindly apply standard rate-distortion metrics to compress the sensor observations into messages to be sent to the agent. The observer's policy is then independent of the actor's task, and can be computed separately.

However, the observation may contain much more information than needed: for example, an image showing an object might be reduced to its coordinates, without losing any meaningful information for the controller. Lossily compressing the message to preserve the relevant information, removing redundant or irrelevant details, can beat the Shannon bound, easing communication requirements without losing performance. Compression is then a projection of the raw observations into a significantly smaller \emph{semantic space}.

As an example, in the CartPole problem analyzed in this work (see Section~\ref{sec:results}), one sensor observation is given by two consecutive 2D camera acquisitions, which only implicitly capture the needed physical system information, e.g., the angular position and velocity of the pole. By employing semantic source coding, the sensor can extract the underlying physical quantities from the raw observations, which can be useful for multiple applications at the receiver, considering not reconstruction accuracy but rather semantic fidelity, which measures the error on the underlying physical values rather than on the raw sensory observation, thus solving the Level B communication problem. What can be gained with the semantic formulation in terms of trade-off between size of the message and performance on the downstream task at the receiver, with respect to the Level A system, is the second question we try to answer with our analysis.

Going further, one may only want to convey the physical information related to a specific task, constructing a Level C system as envisioned in~\cite{shannon1949mathematical}. For this purpose, the sensor has access to feedback signals indicating the quality of the actions of the remote agent when decoding specific messages. This data can be used by the sensor to adjust the observations coding scheme to optimally trade off the size of the message and the performance of the remote agent on the specific control problem. How to design this scheme, and how this scheme compares with Level A and B designed systems, are the other questions we answer with our study.

\subsection{The Information Bottleneck}
\label{subsec:info_bottleneck}
We now consider the observer's encoding function. As the actor can use the received message $m_t$, along with its memory of past messages at previous steps, to choose an action $a_t$, the encoding function is $\Lambda: \Omega\times\mc{M}^{\tau} \rightarrow \mc{M}$, which considers a case with memory $\tau$. The function will generate a message $M_t = \Lambda(O_t,M_{t-\tau},\ldots,M_{t-1})$ to be sent to the actor at each step $t$. As mentioned above, the encoding function $\Lambda$ can be optimized to solve one of the three communication problems.

We assume that the underlying state of the physical system at time $t$ can be entirely described by the variable $S_t$, which is an element of the physical state space $\Sigma$. However, in our remote \gls{pomdp} formulation, the sensor has access to this state only through the observation $O_t$, which is in general a stochastic function of $S_t$ whose codomain is the observation space $\Omega$. In addition, we assume that the performance at time $t$ in a particular task $\mc{T}$
%(i.e., the parameter that the two agents aim to maximize)
depends only on a sufficient statistic $i_{\mc{T}}(s)$ of any given state $s\in\Sigma$. Denoting the number of bits required to represent a realization of a random variable $X$~\cite{cover:IT} as $b(X)$, we assume that
\begin{align}
\label{eq:relation}
    b(i_{\mc{T}}(S)) < b(S) < b(O) \quad \forall S \in \Sigma.
\end{align}
We can observe that ${i_{\mc{T}}(S) \rightarrow S \rightarrow O}$ is a hidden Markov model, as the information available in the current observation also summarizes the full history of past observations.
The random quantity $i_{\mc{T}}(S)$ represents the minimal description of the system with respect to the task $\mc{T}$, i.e., no additional data computed from $S$ adds meaningful information for the actor when solving $\mc{T}$. The state $S$ may also include task-irrelevant physical information on the system. However, both $S$ and $i_{\mc{T}}(S)$ are unknown quantities, as the observer only receives a noisy and high-dimensional representation of $S$ through $O$. This is a well-known issue in \gls{drl}: in the original paper presenting the \gls{dqn} architecture~\cite{mnih2015human}, the agent could only observe the screen while playing classic arcade videogames, and did not have access to the much more compact and precise internal state representation of the game. Introducing communication and dynamic encoding adds another layer of complexity.

\subsection{Encoding and Distortion}\label{subsec:coding}
In our remote \gls{pomdp}, the communication channel between the sensor and the actor is not ideal, thus the information the observer can convey to the actor at each step $t$ is limited to a maximum length of $L$ bits per message. Consequently, the problem introduces an information bottleneck between the observation $O_t$ and the final action $A_t$ the actor can take, which is based on the message $M_t$ conveyed through the channel. In principle, whenever $b(O)>L$, the constrained communication introduces a distortion $d(O, \hat{O})$, whose theoretical asymptotic limits are given by rate-distortion theory~\cite{cover:IT}. If we also consider \emph{memory}, i.e., the use of past messages in the estimation of $\hat{O}$, the mutual information between $O$ and the previous messages can be used to reduce the distortion, improving the quality of the estimate.
In the semantic problem, the aim is to extrapolate the real physical state of the system $S_t$ from the observation $O_t$, which can be a complex stochastic function. In general, the real state lies in a low-dimensional semantic space $\Sigma$. The term semantic is motivated by fact that, in this case, the observer is not just transmitting pure sensory data, but some meaningful physical information about the system. Consequently, the distortion to be considered in this case is some fidelity function $d(S_t, \hat{S}_t)$ measuring the reconstruction quality of the state $S_t$ in the semantic space, where the estimation $\hat{S}_t$ is performed by the actor based on its knowledge of the message, as well as its memory of past messages. To be even more efficient and specific with respect to the task $\mc{T}$, the observer may optimize the message $M_t$ to minimize the distortion
%$d(i_{\mc{T}}(S_t)|M_{t-\tau},\ldots,M_t)$
between the effective representation $i_{\mc{T}}(S_t)$, which contains only the task-specific information, and the knowledge available to the actor. Naturally, any message instance $m_t\in\mc{M}$ must be at most $L$ bits long, in order to respect the constraint. If we define a generic distortion
%$d_{\alpha}(S_t,M_t)$,
$d_{\alpha}$ between the real quantity and its reconstruction at the different levels, which are indexed by $\alpha\in\{A,B,C\}$, the observer's task %\gls{pomdp}
is then defined as follows:
\begin{equation}
\Lambda^*=\argmax_{\mathclap{\Lambda:\Omega\times\mc{M}^{\tau}\rightarrow\mc{M}}}\E{\sum_{t=0}^\infty \gamma^t\left[d_{\alpha} -\beta\ell(M_t)\right]\bigg| S_0,\Lambda},\label{eq:varcapproblem}
\end{equation}
where $\beta\geq0$ is a communication cost parameter and the function $\ell(\cdot)$ represents the length (in bits) of a message. The observation space of the observer is given by $\Omega\times\mc{M}^{\tau}$, while the observation space of the actor is $\mc{M}^{\tau+1}$, as it only receives the messages from the observer.

\section{Proposed Solution}
\label{sec:solution}
 In this section, we introduce the architecture we used to represent $\Lambda$, the \gls{vqvae}, and discuss the remote \gls{pomdp} solution. As the \gls{vqvae} model is not adaptive, we consider an ensemble model with different quantization levels, limiting the choice of the observer to \emph{which} \gls{vqvae} model to use in the transmission. As the action space in~\eqref{eq:varcapproblem} is huge, techniques such as emergent communication that learn it explicitly are limited to scenarios with very simple tasks and immediate rewards. By restricting the problem to a smaller action space, we may find a slightly suboptimal solution, but we can deal with much more complex problems.

\subsection{Deep VQ-VAE Encoding}
\label{subsec:vae}
In order to represent the encoding function $\Lambda$, and to restrict the problem in~\eqref{eq:varcapproblem} to a more manageable action space, the observer exploits the \gls{vqvae} architecture introduced in~\cite{van2017neural}. The \gls{vqvae} is built on top of the more common \gls{vae} model, with the additional feature of finding an optimal discrete representation of the latent space. The \gls{vae} is used to reduce the dimensionality of an input vector $X \in \mathbb{R}^I$, by mapping it into a stochastic latent representation $Z \in \mathbb{R}^L \sim q_{\phi}(Z | X)$, where $L < I$. The stochastic encoding function $q_{\phi}(Z | X)$ is a parameterized probability distribution represented by a neural network with parameter vector $\phi$. To find optimal latent representations $Z$, the \gls{vae} jointly optimizes a decoding function $p_{\theta}(\hat{X}|Z)$ that aims to reconstruct $X$ from a sample $\hat{X} \sim p_{\theta}(\hat{X} | Z)$. This way, the parameter vectors $\phi$ and $\theta$ are usually jointly optimized to minimize the distortion $d(X, \hat{X})$ between the input and its reconstruction, given the constraint on $Z$, while reducing the distance between $q_{\phi}(Z|X)$, and some prior $q(Z)$~\cite{vae2013} used to impose some structure or complexity budget.

However, in practical scenarios, one needs to digitally encode the input $X$ into a discrete latent representation. To do this, the \gls{vqvae} quantizes the latent space by using $N$ $K$-dimensional codewords $z_1, \dots, z_N \in \mathbb{R}^K$, forming a dictionary with $N$ entries. Moreover, to better represent 3D inputs, the \gls{vqvae} quantizes the latent representation $Z$ using a set of $F$ blocks, each quantizing one feature $f(X)$ of the input, and chosen from a set of $N$ possible codewords. We denote the set containing all the $N^F$ possible concatenated blocks with $\mathcal{M}(N)$, as it represents the set of all possible messages the observer can use to convey to the actor the information on the observation $O$, by using $F$ discrete $N$-dimensional features. The peculiarity of the \gls{vqvae} architecture is that it jointly optimizes the codewords in $\mathcal{M}(N)$ together with the stochastic encoding and decoding functions $q_{\phi}$ and $p_{\theta}$, instead of simply applying fixed vector quantization on top of learned continuous latent variables $Z$. When the communication budget is fixed, i.e., the value of $L$ is constant, the protocol to solve the remote \gls{pomdp} is rather simple: first, the observer trains the \gls{vqvae} with $N = 2^{\frac{L}{F}}$ to minimize the technical, semantic, or effective distortion $d_{\alpha}$, depending on the problem; then, at each step $t$, the observer computes $\hat{m} \sim q_{\phi}(\cdot | o_t)$, and finds $ m_t = \argmin_{m \in \mathcal{M}(N)} \| m - \hat{m} \|_2$.
The message $m_t$ is sent to the actor, which can optimize its decision on the action to take accordingly.

\subsection{Dynamic Feature Compression}
\label{subsec:drl}

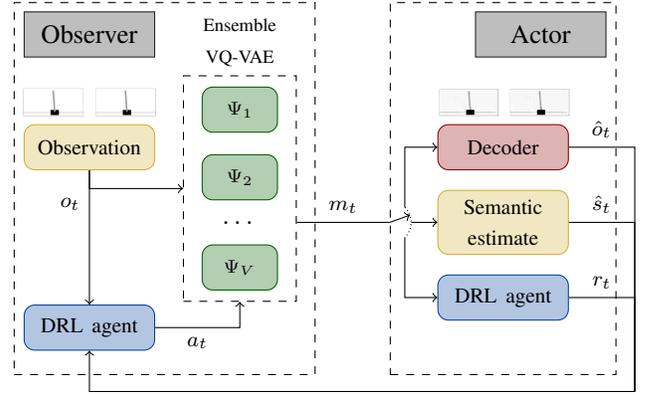
\begin{figure}
\centering
\input{figures/architecture}
\caption{Dynamic feature compression architecture.}\vspace{-0.5cm}
\label{fig:arch}
\end{figure}

We can then consider the architecture shown in Fig.~\ref{fig:arch}, consisting of a set of \glspl{vqvae} $\mathcal{V} = \{ \Psi_1, \ldots, \Psi_V\}$, where each \gls{vqvae} has a different compression level $N_{\Psi}$: as we only consider the communication side of the problem, the actor is trained beforehand using the messages with the finest-grained quantization, which are compressed with the \gls{vqvae} $\Psi$ with larger value of $N$, i.e., larger codebook.
 %with the highest $\Psi_v$. T
 The actor can then perform three different tasks, corresponding to the three communication problems: it can decode the observation (Level A) with the highest possible accuracy, using the decoder part of the \gls{vqvae} architecture; it can estimate the hidden state (Level B) using a supervised learning solution; or it can perform a control action based on the received information and observe its effects (Level C).

In all three cases, the dynamic compression is performed by the observer, based on the feedback from the actor. The observer side of the remote \gls{pomdp}, whose reward is given in~\eqref{eq:varcapproblem}, is restricted to the choice of $\Psi_v$, i.e., to selecting one of the possible codebooks learned by each \gls{vqvae} in the ensemble model. Naturally, the feedback depends on the communication problem that the observer is trying to solve: at Levels A and B, the observer aims at minimizing distortion in the observation and semantic space, respectively, while at Level C, the objective is to maximize the reward of the actor.

In all three cases, \emph{memory} is important: representing snapshots of the physical system in consecutive instants, subsequent observations have high correlations, and the actor can glean a significant amount of information from past messages. This is a significant advantage of dynamic compression, as it can adapt messages to the estimated knowledge of the receiver.

While the observer is adapting its transmissions to the actor's task, the actor's algorithms are fixed. They could themselves be adapted to the dynamic compression strategy, but this joint training is significantly more complex, and we consider it as a possible extension of this work.

\section{Numerical Evaluation}
\label{sec:results}

The underlying use case analyzed in this work is the well-known CartPole problem, as implemented in the \emph{OpenAI Gym} library.\footnote{\url{https://www.gymlibrary.dev/environments/classic_control/cart_pole/}} In this problem, a pole is installed in a cart, and the task is to control the cart position and velocity to keep the pole in equilibrium. The physical state of the system is fully described by the cart position $x_t$ and velocity $\dot{x}_t$, and the pole angle $\psi_t$ and angular velocity $\dot{\psi}_t$. Consequently, the true state of the system is ${s_t = ( x_t, \dot{x}_t, \psi_t, \dot{\psi}_t )}$, and the semantic state space is $\Sigma\subset\mathbb{R}^4$ (because of physical constraints, the range of each value does not actually span the whole real line).

\begin{figure}[t]
\centering
\includegraphics[width=0.8\columnwidth]{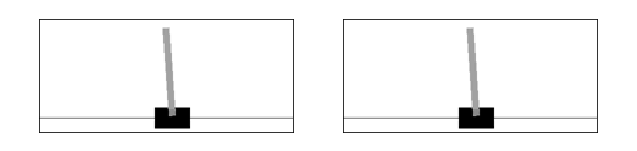}
\includegraphics[width=0.8\columnwidth]{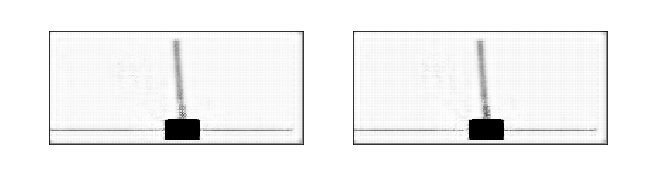}
\caption{Example of the original and reconstructed observation.}\vspace{-0.6cm}
\label{fig:cartpole}
\end{figure}

At each step $t$, the observer senses the system by taking a color picture of the scene, which is in a space $\mc{P}=\left\{ 0, \dots, 255\right\}^{180 \times 360 \times 3}$. To take the temporal element into account, an observation $O_t$ includes two subsequent pictures, at times $t-1$ and $t$, so that the observation space is $\mc{O}=\mc{P}\times\mc{P}$. An example of the transmission process is given in Fig.~\ref{fig:cartpole}, which shows the original version sensed by the observer (above) and the reconstructed version at the receiver (below) when using a trained \gls{vqvae} with $N_{\Psi}=6$.

In the CartPole problem, the action space $\mathcal{A}$ contains just two actions $\mathtt{Left}$ and $\mathtt{Right}$, which push the cart to the left or to the right, respectively. At the end of each step, depending on the true state $s_t$, and on the taken action $a_t$, the environment will return a deterministic reward $ R_t = -x_{\text{max}}^{-1} |x_t|-\psi_{\text{max}}^{-1} |\psi_t|, \  \forall t$,
where $x_{\text{max}} = 4.8$ [m] and $\psi_{\text{max}} = 24^{\circ}$ are the maximal achievable values for the two quantities. The goal for the two agents is thus to maximize the cumulative discounted sum of the reward $R_t$, while limiting the communication cost.

\subsection{The Coding \& Decoding Functions}
\label{subsec:training_vqvae}

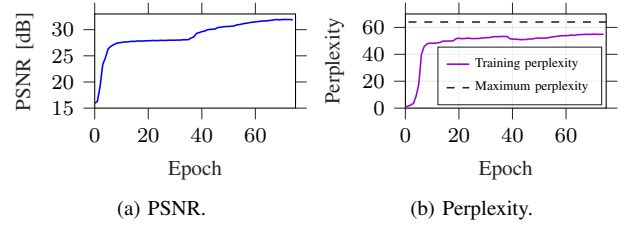
\begin{figure}[t]
\centering
\subfloat[PSNR.\label{fig:vqvae}]{\input{figures/training_vqvae}}%
\subfloat[Perplexity.\label{fig:perplexity}]{\input{figures/perplexity_vqvae}}%
\caption{Training of the \gls{vqvae} model with $N=6$.}\vspace{-0.5cm}
\label{fig:vqvae_training}
\end{figure}

As mentioned in Sec.~\ref{subsec:coding}, the observer can optimize its coding function $\Lambda$ according to different criteria depending on the considered communication problem. However, as we explained in Sec.~\ref{sec:solution}, optimizing the choice of $\Lambda$ is usually not feasible due to the curse of dimensionality on the action space. Consequently, we rely on a pre-trained set $\mc{V}$ of \gls{vqvae} models, whose codebooks are optimized to solve the technical problem, i.e., minimizing the distortion on the observation measured using the \gls{mse}: $d_{A}(O, \hat{O}) = \text{MSE}(O, \hat{O})$. The training performance of the \gls{vqvae} with $N_{\Psi}=6$ is shown in Fig.~\ref{fig:vqvae_training}: the encoder converges to a good reconstruction performance, which can be measured by its \emph{perplexity}. The perplexity is simply $2^{H(p)}$, where $H(p)$ is the entropy of the codeword selection frequency, and a perplexity equal to the number of codewords is the theoretical limit, which is only reached if all codewords are selected with the same probability. The perplexity at convergence is 54.97, which is close to the theoretical limit for a real application.

The observer then uses \gls{drl} to foresightedly optimize the quantization level $\Psi_t$ at each time step, maximizing~\eqref{eq:varcapproblem} for each communication problem. We train the observer to solve the level-specific coding problem by designing three different rewards, depending on the considered communication level:
\begin{enumerate}
\item \emph{Level A (technical problem):} The reward for the observer is $r_t = \text{PSNR}(o_t, \hat{o_t}) - \beta \ell(m_t)$, where the \gls{psnr} is an image fidelity measure proportional to the logarithm of the normalized \gls{mse} between the original and reconstructed image;
\item \emph{Level B (semantic problem):} The reward is $r_t = -\text{MSE}(s_t, \hat{s_t}) - \beta \ell(m_t)$, and the decoder needs to estimate the underlying physical state $s_t$ by minimizing the \gls{mse}, i.e., the distance between $s_t$ and $\hat{s}_t$ in the semantic space.  In our case, the estimator used to obtain $\hat{s}_t$ is a pre-trained supervised \gls{lstm} neural network;
\item \emph{Level C (effective problem):} In this case the reward is $r_t = Q(o_t,\pi(a_t|m_t)) - \beta\ell(m_t)$, where $Q(o_t, a_t)$ is the Q-function of the actor, measuring its long-term performance in the CartPole task when sampling actions $a_t$ according to its policy $\pi$ informed by the message $m_t$. The policy $\pi$ is given by a \gls{dqn} agent implementing an \gls{lstm} architecture, pre-trained using data with the highest available message quality (6 bits per feature).
\end{enumerate}
We observe that, in this case, $i_{\mathcal{T}(s)} = s$, i.e., the task depends on all the semantic features contained in $s_t$. However, the $4$ values do not carry the same amount of information to the actor: depending on the system conditions, i.e., the state $s_t$, some pieces of information are more relevant than others.

\begin{figure*}[t]
\centering
    \subfloat{\input{figures/prob_legend}}\vspace{-0.3cm}\\
        \setcounter{subfigure}{0}
\subfloat[Technical problem.\label{fig:technical}]{\input{figures/tech_prob}}%
\subfloat[Semantic problem.\label{fig:semantic}]{\input{figures/sem_prob}}%
\subfloat[Effective problem.\label{fig:effective}]{\input{figures/eff_prob}}%
\caption{Performance of the communication schemes on the three levels of the remote \gls{pomdp}.}\vspace{-0.5cm}
\label{fig:performance}
\end{figure*}
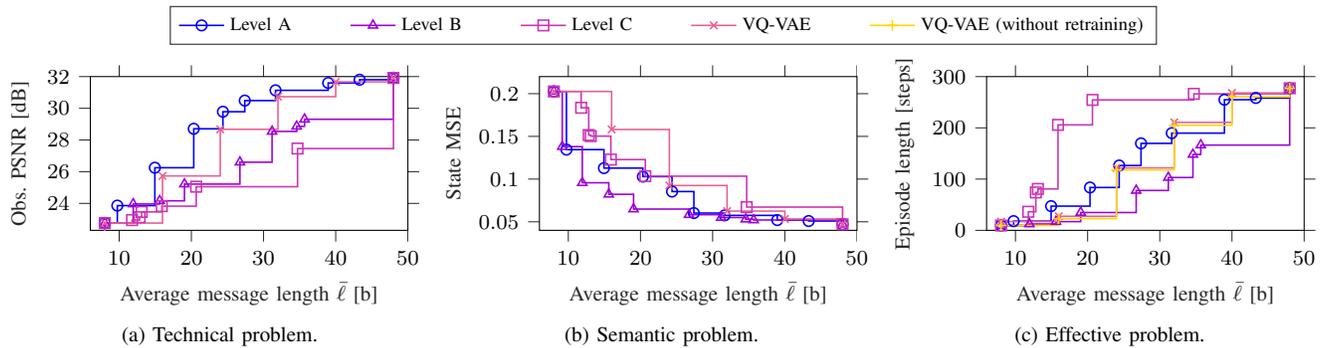

\subsection{Results}
\label{subsec:results}

We assess the performance of the three different tasks in the CartPole scenario by simulation, measuring the results over 100 episodes after convergence. Fig.~\ref{fig:performance} shows the performance of the various schemes over the three problems, compared with a static \gls{vqvae} solution with a constant compression level. In the Level C evaluation, we also consider a static \gls{vqvae} solution in which the actor is not retrained for each $N_{\Psi}$, but is only trained for $N_{\Psi}=6$ (i.e., 48 bits per message). We trained the dynamic schemes with different levels of the communication cost $\beta$, so as to provide a full picture of the adaptation to the trade-off between performance and compression. We also introduce the notion of \emph{Pareto dominance}: an $n$-dimensional tuple $\eta=(\eta_1,\ldots,\eta_n)$, $\eta$ Pareto dominates $\eta'$ (which we denote as $\eta\succ \eta'$) if:
\begin{equation}
\eta\succ \eta' \iff \exists i:\eta_i>\eta_i' \wedge\eta_j\geq\eta_j'\,\forall j.
\end{equation}
We can extend this to different schemes, which have multiple performance points. The definition of Pareto dominance for schemes $x$ and $y$ is: $x\succ y \iff \exists \eta_x\succ\eta_y\,\forall \eta_y.$

We first consider the technical problem performance, shown in Fig.~\ref{fig:technical}: as expected, the Level A dynamic compression outperforms all other solutions, and its performance is Pareto dominant with respect to static compression. Interestingly, the Level B and Level C solutions perform worse than static compression: by concentrating on features in the semantic space or the task space, these solutions remove information that could be useful to reconstruct the full observation, but is meaningless for the specified task.

In the semantic problem, shown in Fig.~\ref{fig:semantic}, a lower \gls{mse} on the reconstructed state is better, and the Level B solution is Pareto dominant with respect to all others. The Level A solution also Pareto dominates static compression, while the Level C solution only outperforms it for higher compression levels, i.e., on the left side of the graph.

Finally, Fig.~\ref{fig:effective} shows the performance at the effectiveness level, summarized by the time that the cart-pole system is within the position and angle limits. The Level C solution significantly outperforms all others, but is not strictly Pareto dominant: when the communication constraint is very tight, setting a static compression and retraining the actor to deal with the specific \gls{vqvae} used may provide a slight performance advantage. However, this is also a possibility for a Level C dynamic compression: retraining the actor to deal with the dynamically compressed messages could match the performance of the retrained static scheme, and in general, almost perfect control can be achieved with less than half of the average bitrate of the static compressor. We also note that, in this case, the Level B solution performs worst: choosing the solution that minimizes the semantic distortion might not be matched to the task, as the state variables have equal weight, while a higher precision might be required when the quantization error might change the action the actor chooses.

\section{Conclusion}
\label{sec:conclusion}

In this work, we presented a dynamic feature compression scheme that can exploit an ensemble \gls{vqvae} to solve the semantic and effective communication problems. The dynamic scheme outperforms fixed quantization, and can be trained automatically with limited feedback, unlike emergent communication models that are unable to deal with complex tasks.

A natural extension of this model is to consider more complex tasks and wider communication channels, corresponding to realistic control scenarios. Another interesting direction for future work is to consider joint training of the actor and the observer, or cases with partial information available at both transmitter and receiver.

\bibliographystyle{IEEEtran}
\bibliography{biblio.bib}

\end{document}

%% file: acronyms.tex
\newacronym{iot}{IoT}{Internet of Things}
\newacronym{ml}{ML}{machine learning}
\newacronym{dl}{DL}{Deep Learning}
\newacronym{marl}{MARL}{multi-agent reinforcement learning}
\newacronym{rl}{RL}{reinforcement learning}
\newacronym{decpomdp}{Dec-POMDP}{Decentralized Partially Observable Markov Decision Process }
\newacronym{pomdp}{POMDP}{Partially Observable Markov Decision Process}
\newacronym{uav}{UAV}{Unmanned Aerial Vehicle}
\newacronym{dqn}{DQN}{Deep Q-Network}
\newacronym{dnn}{DNN}{deep neural network}
\newacronym{dial}{DIAL}{Differentiable Inter-Agent Learning}
\newacronym{mdp}{MDP}{Markov decision process}
\newacronym{fov}{FoV}{Field of View}
\newacronym{cnn}{CNN}{Convolutional Neural Network}
\newacronym{nn}{NN}{neural network}
\newacronym{ddql}{DDQL}{Distributed Deep Q-Learning}
\newacronym{pdf}{PDF}{Probability Density Function}
\newacronym{ndpomdp}{ND-POMDP}{Networked Distributed Partially Observable Markov Decision Process}
\newacronym{radam}{RAdam}{Rectified Adam}
\newacronym{cdf}{CDF}{cumulative distribution function}
\newacronym{mpc}{MPC}{Model Predictive Control}
\newacronym{rv}{rv}{Random Variable}
\newacronym{qoe}{QoE}{Quality of Experience}
\newacronym{tlc}{TLC}{Telecommunications}
\newacronym{cml}{CML}{communications for machine learning}
\newacronym{mlc}{MLC}{machine learning for communications}
\newacronym{drl}{DRL}{Deep Reinforcement Learning}
\newacronym{rf}{RF}{Radio Frequency}
\newacronym{urllc}{URLLC}{Ultra-Reliable and Low-Latency Communications}
\newacronym{fl}{FL}{federated learning}
\newacronym{kpi}{KPI}{Key Performance Indicators}
\newacronym{mec}{MEC}{Mobile Edge Computing}
\newacronym{ei}{EI}{Edge Intelligence}
\newacronym{bs}{BS}{base station}
\newacronym{sdn}{SDN}{Software Defined Networking}
\newacronym{mimo}{MIMO}{Multiple-Input Multiple-Output}
\newacronym{gp}{GP}{Gaussian Process}
\newacronym{iiot}{IIoT}{Industrial Internet of Things}
\newacronym{csi}{CSI}{Channel State Information}
\newacronym{sgd}{SGD}{Stochastic Gradient Descent}
\newacronym{iid}{i.i.d.}{independent and identically distributed}
\newacronym{ofdm}{OFDM}{Orthogonal Frequency Division Multiplexing}
\newacronym{los}{LOS}{Line-of-Sight}
\newacronym{nlos}{NLOS}{Non-Line-of-Sight}
\newacronym{snr}{SNR}{Signal to Noise Ratio}
\newacronym{rb}{RB}{Resource Block}
\newacronym{6g}{6G}{sixth generation}
\newacronym{ai}{AI}{artificial intelligence}
\newacronym{sfl}{SFL}{Synchronous Federated Learning}
\newacronym{frfl}{FRFL}{Fixed Rate Federated Learning}
\newacronym{pgm}{PGM}{Probabilistic Graphical Model}
\newacronym{hmm}{HMM}{Hidden Markov Model}
\newacronym{elbo}{ELBO}{Evidence Lower Bound}
\newacronym{pmf}{PMF}{Probability Mass Function}
\newacronym{smab}{SMAB}{Stochastic Multi-Armed Bandit}
\newacronym{mab}{MAB}{multi-armed bandit}
\newacronym{mc}{MC}{Monte Carlo}
\newacronym{is}{IS}{Importance Sampling}
\newacronym{dms}{DMS}{discrete memoryless source}
\newacronym{ucb}{UCB}{upper confidence bound}
\newacronym{ser}{SER}{Symbol Error Rate}
\newacronym{sc}{SC}{Semantic Communications}
\newacronym{voi}{VoI}{Value of Information}
\newacronym{nlp}{NLP}{natural language processing}
\newacronym{ts}{TS}{Thompson Sampling}
\newacronym{cmab}{CMAB}{contextual multi-armed bandit}
\newacronym{rccmab}{RC-CMAB}{rate-constrained \gls{cmab}}
\newacronym{rcmab}{R-CMAB}{remote \gls{cmab}}
\newacronym{ib}{IB}{information bottleneck}
\newacronym{merl}{MERL}{maximum entropy reinforcement learning}
\newacronym{pac}{PAC}{Probably Approximately Correct}
\newacronym{aoi}{AoI}{Age of Information}
\newacronym{aoii}{AoII}{Age of Incorrect Information}
\newacronym{uoi}{UoI}{Urgency of Information}
\newacronym{vqvae}{VQ-VAE}{Vector Quantized Variational Autoencoder}
\newacronym{vae}{VAE}{Variational Autoencoder}
\newacronym{psnr}{PSNR}{Peak Signal to Noise Ratio}
\newacronym{mse}{MSE}{Mean Square Error}
\newacronym{lstm}{LSTM}{Long Short-Term Memory}

%% file: figures/architecture.tex
\begin{tikzpicture}[every text node part/.style={align=center}]

\node (obsg) at (0,1.6) {\includegraphics[width=2cm]{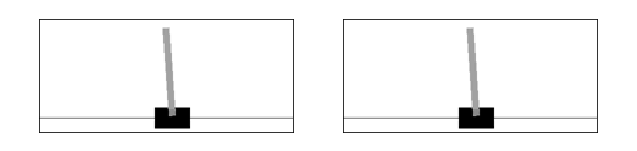}};
\node[draw,rounded corners,draw=gold,fill=gold,fill opacity=0.3,text opacity=1,minimum height=0.6cm,text width=1.5cm,minimum width=1.5cm] (obs) at (0,1) {\footnotesize Observation};

\node[draw,rounded corners,draw=darkblue,fill=darkblue,fill opacity=0.3,text opacity=1,minimum height=0.6cm,text width=1.5cm,minimum width=1.5cm] (odrl) at (0,-1.4) {\footnotesize DRL agent};

\node[text width=1.8cm] (encd) at (2,2.4) {\scriptsize Ensemble VQ-VAE};
\node[draw,dashed,minimum height=3cm,minimum width=1.5cm] (vq) at (2,0.45) {};
\node[draw,rounded corners,draw=darkgreen,fill=darkgreen,fill opacity=0.3,text opacity=1,minimum height=0.6cm,minimum width=1cm] (enc1) at (2,1.5) {\scriptsize $\Psi_1$};
\node[draw,rounded corners,draw=darkgreen,fill=darkgreen,fill opacity=0.3,text opacity=1,minimum height=0.6cm,minimum width=1cm] (enc2) at (2,0.6) {\scriptsize $\Psi_2$};
\node (encd) at (2,0) {$\dots$};
\node[draw,rounded corners,draw=darkgreen,fill=darkgreen,fill opacity=0.3,text opacity=1,minimum height=0.6cm,minimum width=1cm] (encn) at (2,-0.6) {\scriptsize $\Psi_V$};

\node[draw,dashed,minimum height=4.95cm,minimum width=4cm] (observer) at (1,0.45) {};
\node[draw,draw=darkslategray38,fill=darkslategray38,fill opacity=0.3,text opacity=1,minimum height=0.6cm,text width=1.5cm,minimum width=1.5cm] (ch) at (0,2.5) {Observer};

\draw[->] (obs.south) |- (vq.west);
\draw[->] (obs.south) --node[left,near start] {\footnotesize $o_t$} (odrl.north);
\draw[->] (odrl.east) -|node[below,near start] {\footnotesize $a_t$} (vq.south);

\node[draw,rounded corners,draw=darkred,fill=darkred,fill opacity=0.3,text opacity=1,minimum height=0.6cm,text width=1.5cm,minimum width=1.5cm] (dec) at (5.5,1) {\footnotesize Decoder};
\node (obsr) at (5.5,1.6) {\includegraphics[width=2cm]{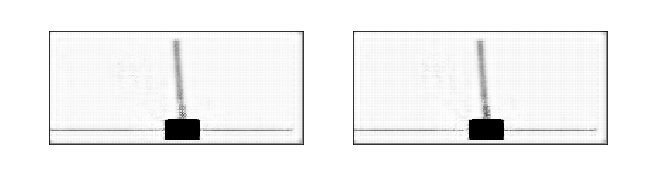}};

\node[draw,rounded corners,draw=gold,fill=gold,fill opacity=0.3,text opacity=1,minimum height=0.6cm,text width=1.5cm,minimum width=1.5cm] (sem) at (5.5,0) {\footnotesize Semantic estimate};

\node[draw,rounded corners,draw=darkblue,fill=darkblue,fill opacity=0.3,text opacity=1,minimum height=0.6cm,text width=1.5cm,minimum width=1.5cm] (adrl) at (5.5,-1) {\footnotesize DRL agent};

\coordinate (corner) at (7.25,-2.25);

\node[draw,dashed,minimum height=4.95cm,minimum width=3cm] (actor) at (5.5,0.45) {};
\node[draw,draw=darkslategray38,fill=darkslategray38,fill opacity=0.3,text opacity=1,minimum height=0.6cm,text width=1.5cm,minimum width=1.5cm] (act) at (6,2.5) {Actor};

\draw[densely dotted] ([xshift=0.2cm,yshift=-0.25cm]actor.west) arc (45:-45:0.285cm);

\draw[-] ([yshift=-0.45cm]vq.east) -- node[above, midway]{\footnotesize $m_t$} ([yshift=-0.45cm]actor.west);
\draw[<-] ([yshift=-0.353cm,xshift=0.268cm]actor.west) -- ([yshift=-0.45cm]actor.west);
\draw[->] ([xshift=0.2cm,yshift=-0.25cm]actor.west) |- (dec.west);
\draw[->] ([xshift=0.285cm,yshift=-0.45cm]actor.west) |- (sem.west);
\draw[->] ([xshift=0.2cm,yshift=-0.65cm]actor.west) |- (adrl.west);

\draw[-] (dec.east) -| node[above, near start]{\footnotesize$\hat{o}_t$} (corner);
\draw[-] (sem.east) -| node[above, near start]{\footnotesize$\hat{s}_t$} (corner);
\draw[-] (adrl.east) -| node[above, near start]{\footnotesize$r_t$} (corner);

\draw[->] (corner) -| (odrl.south);

\end{tikzpicture}

%% file: figures/training_vqvae.tex
% This file was created with tikzplotlib v0.10.1.
\begin{tikzpicture}

\definecolor{darkslategray38}{RGB}{38,38,38}

\begin{axis}[
width=\sfwidth,
height=\sfheight,
tick align=outside,
x grid style={white},
xlabel=\textcolor{darkslategray38}{\footnotesize Epoch},
xmajorgrids,
xmin=0, xmax=75,
xtick style={color=darkslategray38},
y grid style={white},
ylabel=\textcolor{darkslategray38}{\footnotesize PSNR [dB]},
ymajorgrids,
ymin=15, ymax=33,
ytick style={color=darkslategray38}
]
\addplot [semithick, color0]
table {%
0 15.9226927197366
1 16.2911698322239
2 19.1328433605512
3 23.3535802444387
4 24.6218090492673
5 26.3264407897398
6 26.7985371388895
7 27.0996538863748
8 27.3754891026957
9 27.4957999769111
10 27.5448733218585
11 27.6700388960785
12 27.6447155309245
13 27.7211329538633
14 27.7728352885242
15 27.7469071827414
16 27.851561519523
17 27.8251605578609
18 27.8251605578609
19 27.8781239559604
20 27.8781239559604
21 27.9048498545737
22 27.9317412396815
23 27.9048498545737
24 27.9317412396815
25 27.9588001734407
26 27.9317412396815
27 27.9588001734407
28 27.9860287567955
29 27.9860287567955
30 27.9860287567955
31 28.0134291304558
32 28.0134291304558
33 28.0687540164554
34 28.0687540164554
35 28.0966830182971
36 28.4163750790475
37 28.6012091359876
38 29.2811799269387
39 29.4309514866353
40 29.6257350205938
41 29.7061622231479
42 29.9567862621736
43 30.0436480540245
44 30.0877392430751
45 30.1322826573376
46 30.4095860767891
47 30.4575749056068
48 30.5551732784983
49 30.6048074738138
50 30.6048074738138
51 30.6550154875643
52 30.7058107428571
53 30.9151498112135
54 30.9691001300806
55 31.0790539730952
56 31.1918640771921
57 31.249387366083
58 31.3667713987954
59 31.4266750356873
60 31.4874165128092
61 31.5490195998574
62 31.6115090926274
63 31.6749108729376
64 31.6749108729376
65 31.7392519729917
66 31.8045606445813
67 31.8708664335714
68 31.9382002601611
69 31.8708664335714
70 31.9382002601611
71 31.9382002601611
72 31.9382002601611
73 31.9382002601611
74 31.8708664335714
};
\end{axis}
\end{tikzpicture}

%% file: figures/perplexity_vqvae.tex
\begin{tikzpicture}

\pgfplotsset{
compat=1.16,
legend image code/.code={
\draw[mark repeat=2,mark phase=2]
plot coordinates {
(0cm,0cm)
(0.15cm,0cm)        %% default is (0.3cm,0cm)
(0.3cm,0cm)         %% default is (0.6cm,0cm)
};%
}
}

\definecolor{darkslategray38}{RGB}{38,38,38}

\begin{axis}[
width=\sfwidth,
height=\sfheight,
legend cell align={left},
legend style={
  fill opacity=0.4,
  font=\tiny,
  draw opacity=1,
  text opacity=1,
  at={(0.98,0.03)},
  anchor=south east,
  draw=darkslategray38,
},
tick align=outside,
xlabel=\textcolor{darkslategray38}{\footnotesize Epoch},
xmajorgrids,
xmin=0, xmax=75,
ylabel=\textcolor{darkslategray38}{\footnotesize Perplexity},
xtick style={color=darkslategray38},
ymajorgrids,
ymin=0, ymax=70,
ytick style={color=darkslategray38}
]
\addplot [semithick, color1]
table {%
0 1.04485
1 1.46484
2 2.27705
3 3.50883
4 8.96742
5 17.61955
6 40.04723
7 45.71285
8 47.61232
9 48.24294
10 48.25072
11 48.29355
12 48.45544
13 48.88806
14 49.79154
15 49.79394
16 49.8776
17 49.86376
18 50.1734
19 51.74464
20 51.84748
21 51.68901
22 51.70829
23 52.01357
24 51.87248
25 51.63717
26 51.75578
27 51.73644
28 51.93943
29 52.13296
30 52.31753
31 52.42508
32 52.61686
33 53.17795
34 53.07103
35 53.12925
36 53.32611
37 53.21506
38 53.29846
39 51.62523
40 51.33761
41 51.29746
42 51.11416
43 50.86843
44 50.98848
45 50.9832
46 51.36984
47 51.35302
48 51.63236
49 52.13667
50 51.97433
51 52.11483
52 51.99107
53 52.19568
54 52.53453
55 52.97333
56 53.21401
57 53.42107
58 53.74551
59 53.85097
60 53.86063
61 53.91586
62 54.38657
63 54.07111
64 54.51436
65 54.60176
66 54.68778
67 54.90974
68 54.86385
69 54.8196
70 54.96458
71 54.9648
72 54.86458
73 54.8658
74 54.9648
};
\addlegendentry{Training perplexity}
\addplot [semithick, darkslategray38, dashed]
table {%
-4.75 64
77.75 64
};
\addlegendentry{Maximum perplexity}
\end{axis}
\end{tikzpicture}

%% file: figures/prob_legend.tex
% This file was created with tikzplotlib v0.10.1.
\begin{tikzpicture}

\begin{axis}[
    width=0cm,
    height=0cm,
    axis line style={draw=none},
    tick style={draw=none},
    at={(0,0)},
    scale only axis,
    xmin=0,
    xmax=0,
    xtick={},
    ymin=0,
    ymax=0,
    ytick={},
    axis background/.style={fill=white},
    legend style={legend cell align=center, fill opacity=1, align=center, draw=darkslategray38, font=\scriptsize, at={(0, 0)}, anchor=center, /tikz/every even column/.append style={column sep=2em}},
    legend columns=5,
]
\addplot [semithick, color0, mark=o, const plot mark left]
table {%
0 0
};
\addlegendentry{Level A}
\addplot [semithick, color1, mark=triangle, const plot mark left]
table {%
0 0
};
\addlegendentry{Level B}
\addplot [semithick, color2, mark=square, const plot mark left]
table {%
0 0
};
\addlegendentry{Level C}
\addplot [semithick, color3, mark=x, const plot mark left]
table {%
0 0
};
\addlegendentry{VQ-VAE}
\addplot [semithick, color4, mark=+, const plot mark left]
table {%
0 0
};
\addlegendentry{VQ-VAE (without retraining)}

\end{axis}

\end{tikzpicture}

%% file: figures/tech_prob.tex
% This file was created with tikzplotlib v0.10.1.
\begin{tikzpicture}

\definecolor{darkslategray38}{RGB}{38,38,38}

\begin{axis}[
width=\fwidth,
height=\fheight,
tick align=outside,
x grid style={white},
xlabel=\textcolor{darkslategray38}{\footnotesize Average message length $\bar{\ell}$ [b]},
xmajorgrids,
xmin=6, xmax=50,
xtick style={color=darkslategray38},
y grid style={white},
ylabel=\textcolor{darkslategray38}{\footnotesize Obs. PSNR [dB]},
ymajorgrids,
ymin=22.300735, ymax=32,
ytick style={color=darkslategray38}
]
\addplot [semithick, color0, mark=o, const plot mark left]
table {%
8 0
8 22.7586
9.743 23.86
14.94 26.25
20.329 28.7065
24.3734 29.783
27.3993 30.4854
31.6596 31.1333
38.955 31.5997
43.3063 31.8023
48 31.9159
};
\addplot [semithick, color1, mark=triangle, const plot mark left]
table {%
8 0
8 22.7586
11.939 23.9631
15.6087 24.1627
19.0424 25.21831
26.7286 26.59871
31.1811 28.5410857
34.591 28.8609
35.6792 29.307112
48 31.9159
};
\addplot [semithick, color2, mark=square, const plot mark left]
table {%
8 0
8 22.7586
11.7827 22.9468
12.824 23.1294
13.1104 23.4629
15.8968 23.8256
20.6696 25.0539
34.712 27.4593
48 31.9159
};
\addplot [semithick, color3, mark=x, const plot mark left]
table {%
8 0
8 22.7586
16 25.7318
24 28.674
32 30.7332
40 31.6663
48 31.9159
};
\end{axis}

\end{tikzpicture}

%% file: figures/sem_prob.tex
% This file was created with tikzplotlib v0.10.1.

\begin{tikzpicture}

\definecolor{darkslategray38}{RGB}{38,38,38}

\begin{axis}[
width=\fwidth,
height=\fheight,
tick align=outside,
x grid style={white},
xlabel=\textcolor{darkslategray38}{\footnotesize Average message length $\bar{\ell}$ [b]},
xmajorgrids,
xmin=6, xmax=50,
xtick style={color=darkslategray38},
y grid style={white},
ylabel=\textcolor{darkslategray38}{\footnotesize State MSE},
ymajorgrids,
yticklabel style={/pgf/number format/fixed},  
ymin=0.04, ymax=0.22,
ytick style={color=darkslategray38}
]
\addplot [semithick, color0, mark=o, const plot mark left]
table {%
8 0.2026
9.743 0.1345
14.94 0.11288
20.329 0.10288
24.3734 0.08542
27.3993 0.06023
31.6596 0.05749
38.955 0.05201
43.3063 0.0509
48 0.0471
};
\addplot [semithick, color1, mark=triangle, const plot mark left]
table {%
8 0.2026
9.1619 0.1380
11.939 0.0957
15.6087 0.0822
19.0424 0.0649
26.7286 0.0583
31.1811 0.055
34.591 0.053
35.6792 0.0521
48 0.0471
};
\addplot [semithick, color2, mark=square, const plot mark left]
table {%
8 0.2026
11.7827 0.1835
12.824 0.1516
13.1104 0.1503
15.8968 0.12286
20.6696 0.1034
34.712 0.06721
48 0.0471
};
\addplot [semithick, color3, mark=x, const plot mark left]
table {%
8 0.2026
16 0.1581
24 0.0925
32 0.0625
40 0.0533
48 0.0471
};
\end{axis}

\end{tikzpicture}

%% file: figures/eff_prob.tex
% This file was created with tikzplotlib v0.10.1.
\begin{tikzpicture}

\definecolor{darkslategray38}{RGB}{38,38,38}

\begin{axis}[
width=\fwidth,
height=\fheight,
tick align=outside,
x grid style={white},
xlabel=\textcolor{darkslategray38}{\footnotesize Average message length $\bar{\ell}$ [b]},
xmajorgrids,
xmin=6, xmax=50,
xtick style={color=darkslategray38},
y grid style={white},
ylabel=\textcolor{darkslategray38}{\footnotesize Episode length [steps]},
ymajorgrids,
ymin=0, ymax=300,
ytick style={color=darkslategray38}
]
\addplot [semithick, color0, mark=o, const plot mark left]
table {%
8 9.404
9.743 17.94
14.94 47.24
20.329 83.62
24.3734 126.54
27.3993 169.71
31.6596 189.71
38.955 254.91
43.3063 258.33
48 277.242
};
\addplot [semithick, color1, mark=triangle, const plot mark left]
table {%
8 9.404
11.939 12.4
15.6087 17.03
19.0424 34.75
26.7286 77.8
31.1811 103.1
34.591 148.07
35.6792 166.38
48 277.242
};
\addplot [semithick, color2, mark=square, const plot mark left]
table {%
8 9.404
11.7827 36.23
12.824 73.92
13.1104 80.81
15.8968 205.79
20.6696 254.52
34.712 266.396
48 277.242
};
\addplot [semithick, color3, mark=x, const plot mark left]
table {%
8 17.1414141414141
16 27.4848484848485
24 121.808080808081
32 210.686868686869
40 268.111111111111
48 277.242424242424
};
\addplot [semithick, color4, mark=+, const plot mark left]
table {%
8 9.4040404040404
16 22.7676767676768
24 117.808080808081
32 204.979797979798
40 260.494949494949
48 277.242424242424
};
\end{axis}

\end{tikzpicture}